\documentclass[10.5pt,compsoc]{TsT}
\usepackage{graphicx}
\usepackage{footmisc}
\usepackage{subfigure}
\usepackage{url}
\usepackage{multirow}
\usepackage[noadjust]{cite}
\usepackage{amsmath,amsthm}
\usepackage{amssymb,amsfonts}
\usepackage{booktabs}
\usepackage{color}
\usepackage{ccaption}
\usepackage{booktabs}
\usepackage{float}
\usepackage{fancyhdr}
\usepackage{caption}
\usepackage{xcolor,stfloats}
\usepackage{comment}
\graphicspath{{figures/}}
\usepackage{cuted}  %flushend,
\usepackage{captionhack}
\usepackage{epstopdf}
\usepackage{diagbox}%
\headevenname{\normalsize{\textbf{\emph{Tsinghua Science and Technology, xxxxx}} 20xx, 2x(x): xxx--xxx}}%
\headoddname{{\sf Wu et al.:}\quad {\textbf{\emph{New Benchmark for Household Garbage Image Recognition}}}}%
\newtheoremstyle{mystyle}{0pt}{0pt}{\normalfont}{1em}{\bf}{}{1em}{}
\theoremstyle{mystyle}
\renewcommand\figurename{Fig.~}

\makeatletter
\renewcommand{\@biblabel}[1]{[#1]\hfill}
\makeatother

\begin{document}

%\thispagestyle{empty}

%\begin{strip}\zihao{3}
%\noindent
%\\ \textbf{Template for Preparation of Manuscripts for \\ \emph{Tsinghua Science and Technology}}
%\vskip 6mm
%\zihao{5}

%\noindent
%This template is to be used for preparing manuscripts for submission to \emph{Tsinghua Science and Technology}. Use of this template will save time in the review and production processes and will expedite publication. However, use of the template is not a requirement of submission. Do not modify the template in any way (delete spaces, modify font size/line height, etc.).
%\vspace{180mm}
%\end{strip}
%\clearpage

\hyphenpenalty=50000

\makeatletter
\newcommand\mysmall{\@setfontsize\mysmall{7}{9.5}}

%%%%%%%%%%%%%%%%%%%%%%%
\newenvironment{tablehere}
  {\def\@captype{table}}
  {}
\newenvironment{figurehere}
  {\def\@captype{figure}}
  {}
%%%%%%%%%%%%%%%%%%%%%%%
%%%%%%%%%%%%%%%%%%%%%%%%%%%%%%%%

\thispagestyle{plain}%
\thispagestyle{empty}%

\let\temp\footnote
\renewcommand \footnote[1]{\temp{\normalsize #1}}
{}
\vspace*{-40pt}
\noindent{\normalsize\textbf{\scalebox{0.885}[1.0]{\makebox[5.9cm][s]
			{TSINGHUA\, SCIENCE\, AND\, TECHNOLOGY}}}}

\vskip .2mm
{\normalsize
	\textbf{
		\hspace{-5mm}
		\scalebox{1}[1.0]{\makebox[5.6cm][s]{%
				I\hfill S\hfill S\hfill N\hfill{\color{white}%
					l\hfill l\hfill}1\hfill0\hfill0\hfill7\hfill-\hfill0\hfill2\hfill1\hfill4
				\hfill \color{white}{\quad 0\hfill ?\hfill /\hfill ?\hfill ?\quad p\hfill p\hfill  ?\hfill ?\hfill ?\hfill --\hfill ?\hfill ?\hfill ?}\hfill}}}}

\vskip .2mm
{\normalsize
	\textbf{
		\hspace{-5mm}
		\scalebox{1}[1.0]{\makebox[5.6cm][s]{%
				DOI:~\hfill~\hfill1\hfill0\hfill.\hfill2\hfill6\hfill5\hfill9\hfill9\hfill/\hfill T\hfill S\hfill T\hfill.\hfill2\hfill0\hfill 2\hfill 1\hfill.\hfill9\hfill0\hfill1\hfill0\hfill x\hfill x\hfill x}}}}

\vskip .2mm\noindent
{\normalsize\textbf{\scalebox{1}[1.0]{\makebox[5.6cm][s]{%
				\color{black}{V\hfill o\hfill l\hfill u\hfill m\hfill%
					e\hspace{0.356em}xx,\hspace{0.356em}N\hfill u\hfill%
					m\hfill b\hfill e\hfill r\hspace{0.356em}x,\hspace{0.356em}%
					x\hfill x\hfill x\hfill x\hfill x\hfill%
					x\hfill x\hfill \hspace{0.356em}2\hfill0\hfill x\hfill x}}}}}\\

\begin{strip}
{\center \vskip 3mm
{\LARGE\textbf{
New Benchmark for Household Garbage Image Recognition}}
\vskip 9mm}

{\center {\sf \large
Zhize Wu, Huanyi Li, Xiaofeng Wang$^{*}$,  Zijun Wu, Le Zou, Lixiang Xu, and Ming Tan}
\vskip 5mm}

\centering{
\begin{tabular}{p{160mm}}

{\normalsize
\linespread{1.6667} %
\noindent
\bf{Abstract:} {\sf
Household garbage images are usually faced with complex backgrounds, variable illuminations, diverse angles, and changeable shapes, which bring a great difficulty in garbage image classification. 
Due to the ability to discover problem-specific features, deep learning and especially convolutional neural networks (CNNs) have been successfully and widely used for image representation learning.
However, available and stable household garbage datasets are insufficient, which seriously limits the development of research and application. 
Besides, the state of the art in the field of garbage image classification is not entirely clear. 
To solve this problem, in this study, we built a new open benchmark dataset for household garbage image classification by simulating different lightings, backgrounds, angles, and shapes. 
This dataset is named \emph{30 Classes of Household Garbage Images} (\mbox{HGI-30}), which contains  18,000 images of 30 household garbage classes.
%We name the data set 30 Types of Household Garbage Images (HGI-30).
%It consists of 30 types of garbage and 18,000 images, each contains one or more garbage. 
The publicly available \mbox{HGI-30} dataset allows researchers to develop accurate and robust methods for household garbage recognition.
%can provide a baseline for researchers when developing new methods. 
We also conducted experiments and performance analysis of the state-of-the-art deep CNN methods on \mbox{HGI-30}, which serves as baseline results on this benchmark.
}
}
\end{tabular}
}
\vskip 6mm

\vskip -3mm
\small\end{strip}

\thispagestyle{plain}%
\thispagestyle{empty}%
\makeatother
\pagestyle{tstheadings}

\begin{figure}[b]
\vskip -6mm
\begin{tabular}{p{44mm}}
\toprule\\
\end{tabular}
\vskip -4.5mm
\noindent
\setlength{\tabcolsep}{1pt}
\begin{tabular}{p{1.5mm}p{79.5mm}}
%&
$\bullet$& Zhize Wu,  Xiaofeng Wang, Zijun Wu, Le Zou, Lixiang Xu, and Ming Tan are with School of Artificial Intelligence and Big Data, Hefei University, Hefei, 230601, China. E-mail: \{wuzz, wangxf, wuzj, zoule, lixiangxu, and tanming \}@hfuu.edu.cn.\\ 
$\bullet$& Huan-Yi Li is with School of Energy Materials and Chemical Engineering, Hefei University, Hefei 230601, Anhui, China. Email: lihuanyiii@163.com.\\
$\sf{*}$&
To whom correspondence should be addressed. \\
          & Manuscript received: 2021-08-20;  accepted: 2021-09-18

\end{tabular}
\end{figure}\large

\vspace{3.5mm}
\section{Introduction}
\label{s:introduction}
\noindent 
Reasonable garbage management has attracted increasing attention~[1]. 
Garbage is a misplaced resource.
As a key concern in garbage recycling, household garbage classification involves many fields and disciplines, and plays an increasingly important role in environmental protection.

\begin{figure}[tb]%
	\centering%
	\includegraphics[width=0.477\textwidth]{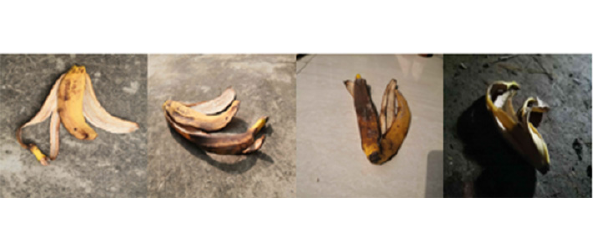}%
	\caption{Changes in shape, color, and light of a banana peel}%
	\label{fig:fig-1}%
\end{figure}

With the advancements of the graphics processing unit (GPU) and computer hardware in recent years, object recognition algorithms based on convolutional neural networks (CNNs) have rapidly developed and been widely used in the environmental protection field~[2,3,4,5]. 
Although excellent performance has been achieved in the garbage classification field, household garbage classification still faces challenges from the large variance in varieties, variable states, and complex textures, which may lead to a significant performance drop.
As shown in Figure~\ref{fig:fig-1},  banana peels have different shapes, and their colors and textures will change with time and temperature. 
Complex backgrounds and varied viewpoints in the real world also bring great difficulties to object recognition. 

%According to the characteristics of garbage, researchers have proposed many excellent algorithms 
Existing approaches~[6,7,8,9] mainly focus on how to quickly and accurately distinguish the categories and find the locations of target garbage in images. 
Many researchers have also proposed garbage classification tools, such as smart trash can~[10,11,12,13]. 

However, these works were all tested and evaluated on \mbox{self-built}, non-public datasets, making it difficult to comprehensively evaluate the performance of algorithms. 
Furthermore, the used datasets are small and simple, which makes convergence difficult when training a deep model.
For example, the dataset used by Rabano et al. in~[14] consisted of only seven  garbage classes and approximately 2,500 images, which are not sufficient for training a normal CNN model.
Therefore, it is necessary to build a public dataset that is validated by \mbox{open-source} algorithms.

Accordingly, in this study, we built an unprecedented dataset with 18,000 household garbage images, i.e., the \emph{30 Classes of Household Garbage Images} (\mbox{HGI-30}) dataset. 
\mbox{HGI-30} contains a total of 30 common household garbage categories. 
The garbage objects in the collected images include complex backgrounds, different illuminations, resolutions, and angles. 
In this paper, we will present the collection, augmentation, label, and evaluation methods of the dataset in detail. 
Furthermore, we provide the evaluation results of the \mbox{state-of-the-art} deep CNN (DCNN) methods, which can be served as baseline results for new methods on this dataset. 
We believe that this benchmark study could provide reference ideas for related fields.

The purpose of this research is to contribute to the fields of garbage recognition and object recognition using CNNs.
\begin{enumerate}%
	\item We built a dataset for garbage recognition and introduced the methods of garbage collection and augmentation, which can extend the research on object recognition to the field of garbage classification. %
	\item The released garbage dataset can reflect the advantages and disadvantages of image classification algorithms, and provide reference and evaluation results for the following researchers in garbage image classification.%
	\item We make the \mbox{HGI-30} dataset available in an online repository~[15].%
\end{enumerate}%

The rest of the article is structured as follows:
In Section 2, we retrospect relevant algorithms and the most advanced technologies in the field of garbage image classification. 
In Section 3, we provide the collection details of the \mbox{HGI-30} dataset.
In Section 4, we validate and evaluate the performance of the state-of-the-art classification methods on the \mbox{HGI-30} data set.
In Section 5, we summarize this research and present future works. 
Finally, we present the method to obtain the HGI-30 dataset from its public repository.

\section{Related Works}
\label{s:related-work}
\noindent
In this section, we mainly review the feature extraction methods and object recognition algorithms, and then introduce the \mbox{state-of-the-art} algorithms in the field of garbage classification.

\subsection{Feature extraction methods based on traditional algorithms}
Feature extraction is the core problem of object recognition. 
It mainly focuses on low-level feature extraction, such as texture, edges, corners, and colors.
The local binary patterns (LBP) descriptor~[16] is the most classic method to extract local texture features by converting the texture into a binary vector. 
In~[17], for instance, Zhang et al. proposed an \mbox{LBP-based} face detection method, which brought about remarkable improvement, especially in terms of the detection accuracy. 
Although the LBP descriptor can quickly compute and is invariant in rotation and gray aspects, it has poor stability in image scale and brightness. 

The \mbox{scale-invariant} feature transform (SIFT)~[18] can overcome such weaknesses in LBP and has certain stability to noise and occlusion. 
This feature has been used for image representation in a variety of scenes.
For instance, in [19], Tao et al. described an airport through the SIFT descriptor. 

The histogram of oriented gradient (HOG)~[20] is another successful feature descriptor, which extracts features based on the histogram of gradient direction. 
Unlike the key point extraction in the SIFT, HOG focuses on extracting edge features. 
By processing the local squares of an image, HOG can maintain good robustness to the geometric and optical deformations. 
It has been consistently applied to various visual analysis tasks. 
For instance, Pang et al.~[21] constructed the HOG features of intersecting detection windows through the reuse of block features, which significantly improved the accuracy of human body detection.

Although these traditional methods have brought about remarkable results to feature extraction, many aspects still need to be improved. 
On the one hand, such traditional algorithms do not work well in complex scenarios. 
On the other hand, it only extracts the low-level semantic features of images and ignores the high-level semantic features, which leads to the loss of several problem-specific features. 
Therefore, high-level semantic features are difficult to be extracted from the object layer using traditional feature extractors, resulting in bottlenecks in the recognition performance.

\subsection{Feature extraction and object recognition algorithm based on CNNs}
Recently, with the development of GPU technology and the research and improvement of large-scale image datasets, deep learning has experienced tremendous development. 
DCNNs are the most widely used deep-learning method.
With the development of DCNNs, the accuracy of image classification on public benchmarks, such as those in~[22,23], has been significantly boosted~[24–27]. 
Because of the excellent ability to extract high-level semantic features~[28,29], DCNNs are also widely used in image generation~[30,31], object recognition~[32,33,34], and object tracking~[35,36].

Compared to traditional algorithms~[37,38], CNN architectures can indeed improve classification performance. 
Mature CNNs, like faster region based CNN ( Faster RCNN)~[39],  Mask RCNN~[40], and deformable convolutional network (DCN)~[41], are outstanding representatives of the application of region proposals, which are often referred to as two-stage algorithms. 
In~[42], for instance, Nie et al. applied Faster RCNN with the backbone of ResNet-50 to detect 3,984 garbage images. 
The results show that the accuracy of garbage recognition is 89.68\%, which is nearly 10 percentage points ahead of the compared traditional approaches.
These two-stage algorithms extract the regions of interest from the input image and classify them. 
Particularly, bringing all the candidate proposal areas into the training improved the accuracy, but the speed was not satisfactory. 
Hence, how to increase speed is a concern for researchers.

Different from the \mbox{two-stage} methods, \mbox{one-stage} methods formulate object detection as a regression problem.
\mbox{Single-shot} detector (SSD)~[43], You Only Look Once (YOLO)~[44-46], M2Det~[47], and EfficientDet~[48] are representative algorithms for \mbox{one-stage} object detection. 
YOLO is probably the most popular one.%
The core of \mbox{one-stage} algorithms is to set a large number of default boxes on each feature map extracted from images. 
We only need to train boxes containing target objects.
This mechanism can reduce the amount of computation and thus improve the speed, but at the cost of a slight drop in accuracy. 
Recent one-stage algorithms have attempted to strike a balance between speed and precision, and they work very well. 
For instance, Chen et al.~[49] proposed an improved YOLOv4 algorithm to detect 15 new types of garbage, with an average accuracy of 64\%.

Many researchers have also performed considerable studies on garbage classification, but the experiments were conducted on \mbox{self-built} datasets. 
The available and stable household garbage datasets are also insufficient, which seriously hinders the development of research and application. 
Moreover, the state of the art in the field of garbage classification and detection is not entirely clear. 
To solve these problems, we built a new open benchmark dataset for household garbage image classification and detection.

\section{Design of the Garbage of Household Dataset}
\label{s:HGI}
\noindent
In this section, the composition and construction details of the dataset are described.
\begin{table}[tb]
	\centering
    \setlength{\belowcaptionskip}{0.2cm}
	\small
	\caption{Numbers of garbage classes in the \mbox{HGI-30} dataset\\}%
	\label{tab:HGI-30}
	\begin{tabular}{l|l|l|l}	
		\hline
		\textbf{Classes} & \textbf{No.} & \textbf{Classes} &\textbf{No.}\\
		\hline
		Applecore& 679 & Paper cup&	579\\
		Banana peel& 622& Pencil&524\\
		Battery	&663& Plastic bottle& 654\\
		Book& 756& Remotecontrol& 536\\
		Buttonbattery& 706& Rice& 559\\
		Can& 687& Shoe&621\\
		Capsule& 702& T-shirt& 564\\
		Carton& 586& Tea leaf& 696\\
		Cigarette butt& 614& Thermometer&553\\
		Cigarette case& 611&Tin can&525\\
		Glass bottle&645&Toothbrush&481\\
		Lunch box&669&Trousers&496\\
		Mask&682&Vegetable leaf&558\\
		Mobilephone& 538& Waste paper&567\\
		Modulator tube& 588& Watermelon peel&517\\		
		\hline
	\end{tabular}
\end{table}

\begin{figure*}[tb]%
	\centering%
	\includegraphics[width=0.977\textwidth]{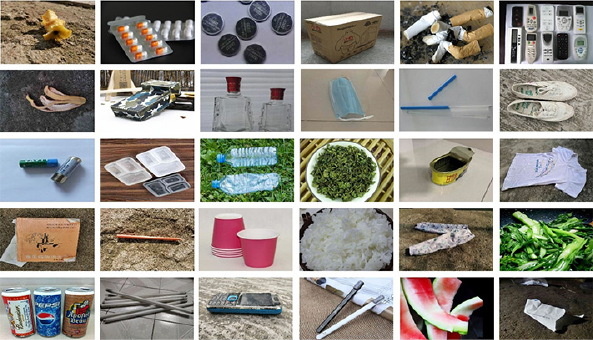}%
	\caption{Samples of each type of garbage in the \mbox{HGI-30} dataset}%
	\label{fig:fig-2}%
\end{figure*}

To promote the application of deep learning in the field of environmental protection and improve the performance of object recognition algorithm for garbage images, we built the \mbox{HGI-30} dataset, which contains 30 household garbage categories, totaling 18,000 images. 
The garbage in the dataset consists of the following characteristics: fixed shape, variable shape, fixed texture, variable texture, and different scales. 
Each image was captured and labeled by experts with professional knowledge of computer vision. 
The unlabeled dataset is used for classification, and the labeled data set is used for detection. 
The number of each category of household garbage and the specific information is presented in Table~\ref{tab:HGI-30}. 
%There are totally 30 kinds of garbage samples 
In Figure~\ref{fig:fig-2}, we illustrate a sample in each category.
%
%\begin{figure}[tb]%
%	\centering%
%	\includegraphics[width=0.477\textwidth]{figure-3.pdf}%
%	\caption{Changes in the sample at different angles, lighting, backgrounds, and resolutions}%
%	\label{fig:fig-3}%
%\end{figure}%
%

Six variations are simultaneously considered in the construction of the \mbox{HGI-30} dataset, namely, viewpoint, background, illumination, resolution, augmentation, and number of objects. 
The objects in the images are also labeled by different people to provide ground truths for evaluation and assessment purposes.

\begin{figure}[tb]%
	\centering%
	\includegraphics[width=0.49\textwidth]{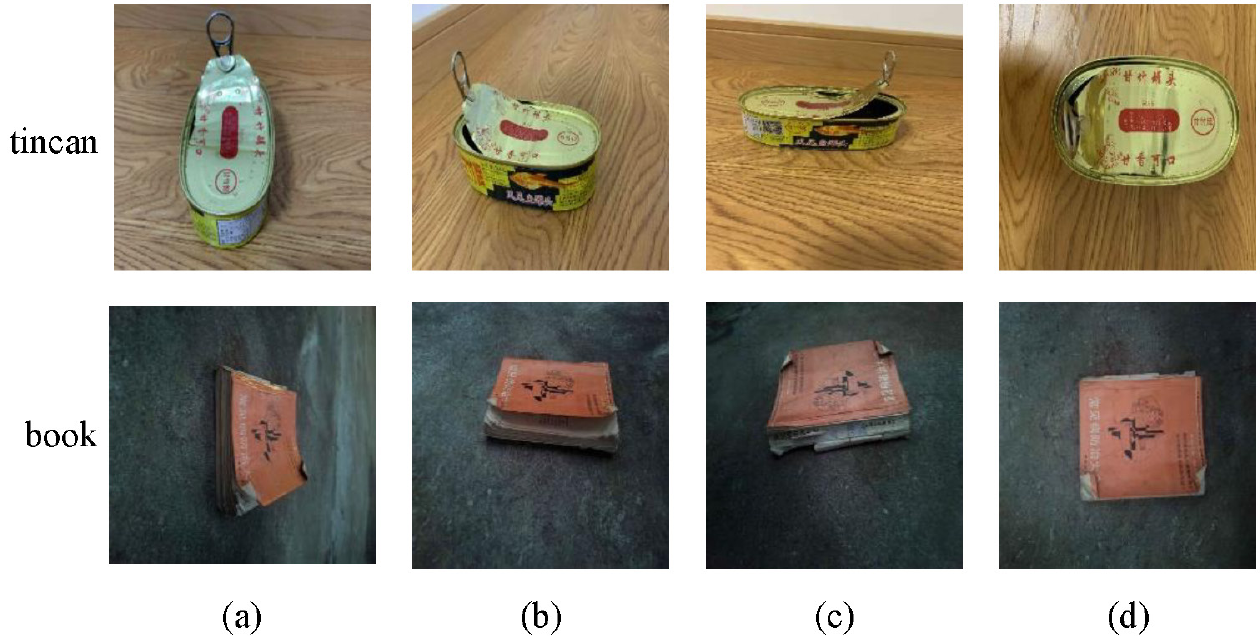}%
	\caption{Two garbage objects at four different viewpoints: (a) front, (b) left, (c) right, and (d) top views of each object}%
	\label{fig:fig-viewpoint}%
\end{figure}

\begin{figure}[tb]%
	\centering%
	\includegraphics[width=0.48\textwidth]{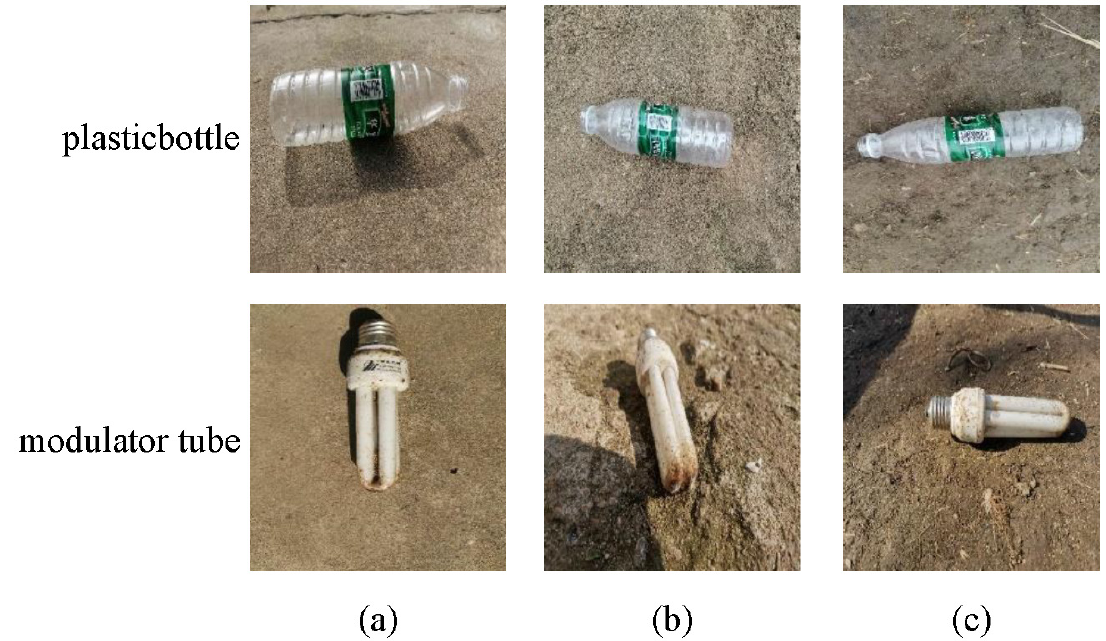}%
	\caption{Two garbage objects at three different backgrounds: (a) cement, (b) sand, and (c) earth}%
	\label{fig:fig-background}%
\end{figure}

\begin{figure}[tb]%
	\centering%
	\includegraphics[width=0.48\textwidth]{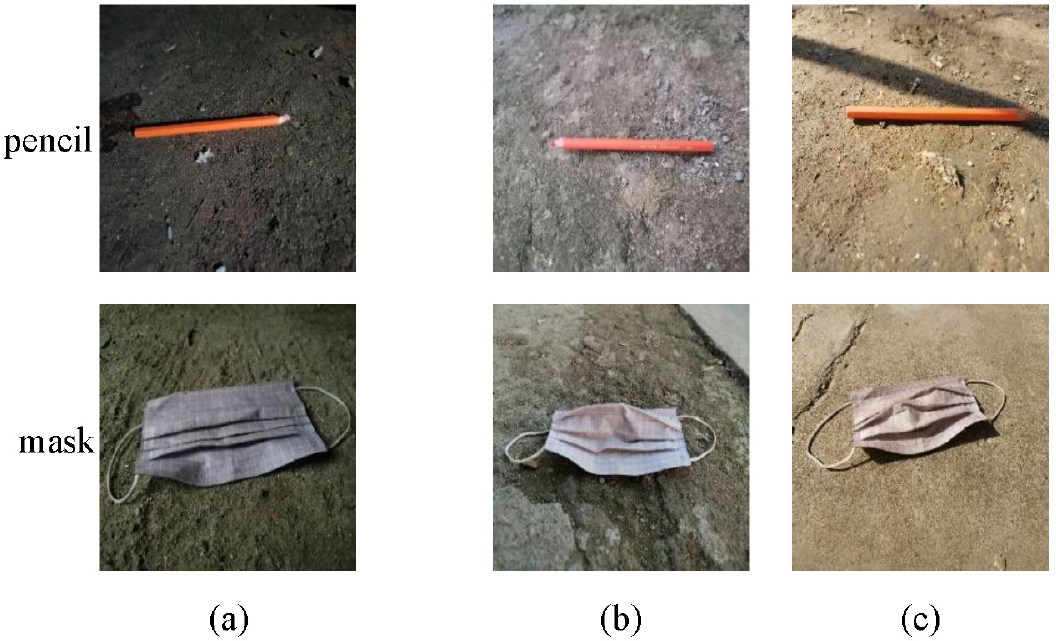}%
	\caption{Two garbage objects at three different lightings: (a) dark, (b) normal, and (c) hard light}%
	\label{fig:fig-lighting}%
\end{figure}

\begin{figure}[tb]%
	\centering%
	\includegraphics[width=0.48\textwidth]{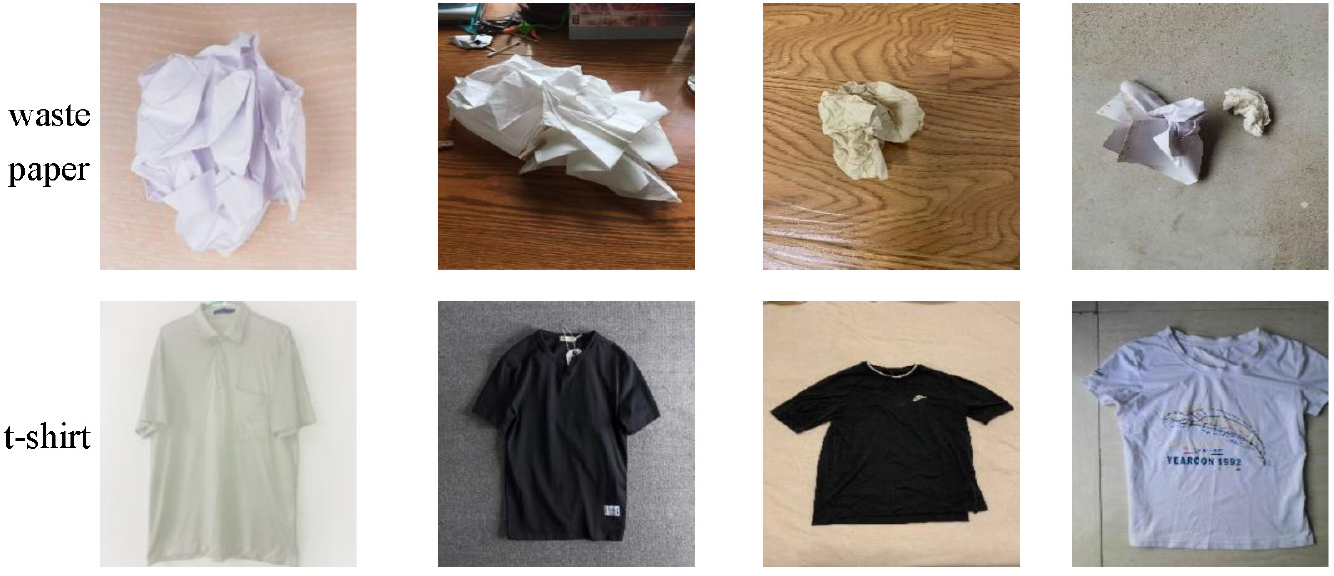}%
	\caption{Two garbage objects at different resolutions, ranging from $300\times400$ to $3000\times4000$}%
	\label{fig:fig-resolution}%
\end{figure}

Specifically, we use four viewpoints, i.e., front, left, right, and top sides, for each garbage category. 
Figure~\ref{fig:fig-viewpoint} shows two garbage objects at four different viewpoints.
There are three different backgrounds for each garbage object, as shown in Figure~\ref{fig:fig-background}.
For the lighting setting, we regularly apply three different settings, i.e., dark, normal, and hard light.
Figure~\ref{fig:fig-lighting} shows two garbage objects at the three different lightings.
For each category, we randomly select different resolutions, ranging from $300\times400$ to $3000\times4000$.
In Figure~\ref{fig:fig-resolution}, we present two garbage objects at different resolutions.

\begin{figure}[tb]%
	\centering%
	\includegraphics[width=0.48\textwidth]{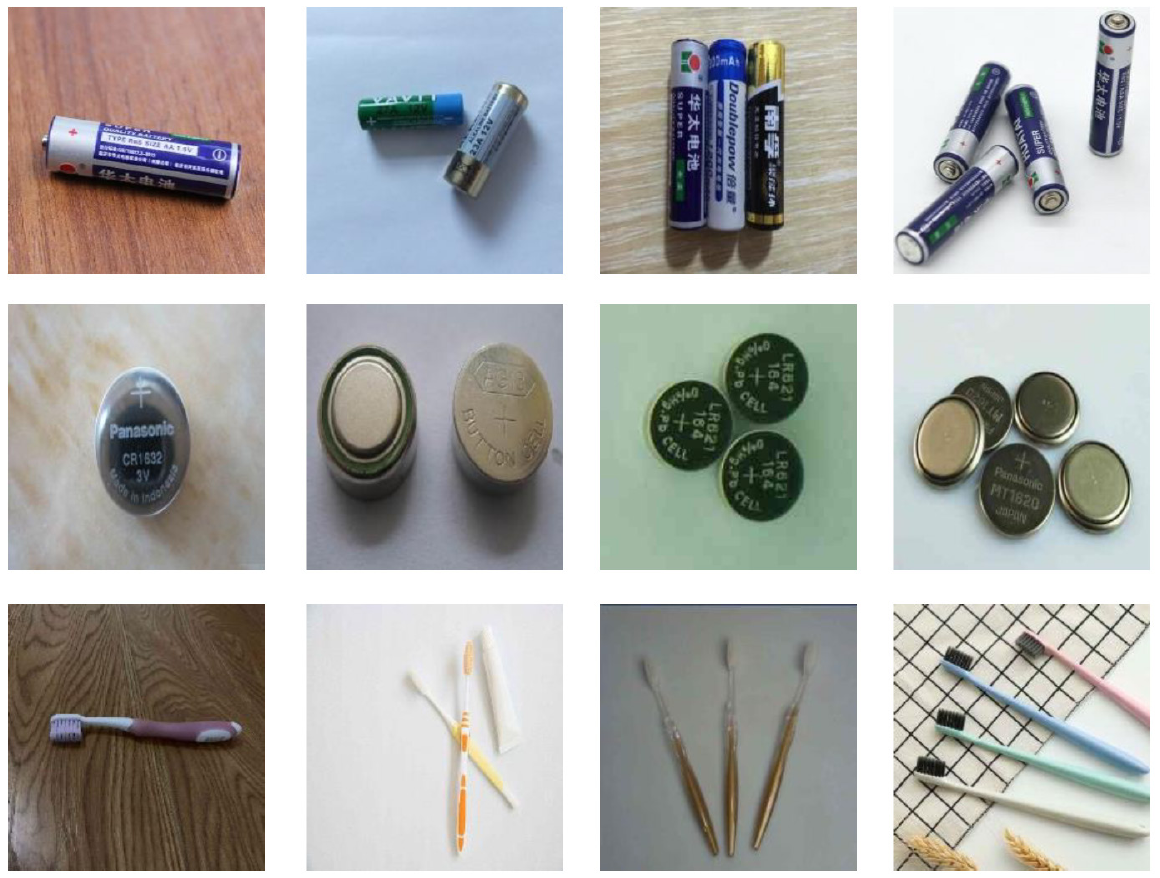}%
	\caption{Three garbage objects with multiple instances}%
	\label{fig:fig-multiple-instance}%
\end{figure}

The number of objects in an image has an important influence on the garbage detection results.
In reality, \mbox{multiple-instance} detection is a more challenging issue than \mbox{single-instance} detection. 
To evaluate the performance of an algorithm on \mbox{multiple-instance} detection, multiple scenarios are included in the \mbox{HGI-30} dataset. 
Figure~\ref{fig:fig-multiple-instance} shows some of the garbage images captured with a single  instance and mutltiple instances.
%
%To better simulate the properties of garbage and the environment around the garbage, different angles were used to simulate the spatial position of the real situation, the brightness and shade were used to simulate the change of illumination. 
%Also, different backgrounds were used to simulate the changing background in the real environment. 
%In addition, various resolution images can be obtained by different acquisition methods. 
%The details are shown in Figure~\ref{fig:fig-3}.
%

\begin{figure}[tb]%
	\centering%
	\includegraphics[width=0.48\textwidth]{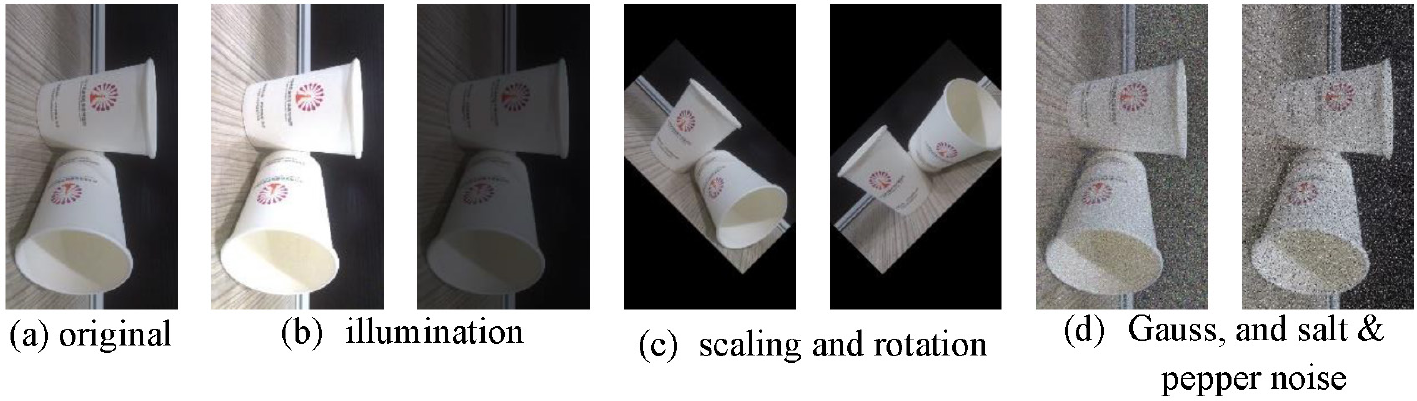}%
	\caption{Three data augmentation methods for objects with few samples}%
	\label{fig:fig-augmentation}%
\end{figure}

CNNs are sensitive to spatial location. 
When objects are changed in the spatial distribution, it misjudges objects into other categories~[50].
Hence, we perform data augmentation on some garbage images that are difficult to collect, such as apple cores and toothbrushes. 
We apply three specific augmentations, namely, rotation, noise, and illumination variation. 
As shown in Figure~\ref{fig:fig-augmentation}, we randomly scale the image with rotation, add Gaussian noise and \mbox{salt-and-pepper} noise, and increase brightness or darkness. 
Such features make this dataset challenging.

\section{Evaluation of State-of-the-Art algorithms on the HGI-30 Dataset}
\label{s:experiment}
\noindent
This section mainly introduces the details of the experiment, including garbage classification and detection on unlabeled and labeled \mbox{HGI-30}. 
The details of each experiment and visual and theoretical analysis of the experimental results are presented separately.

The following experiments were run on a computer with an Intel i7-7700 CPU with 3.6GHz, 16GB RAM, and two NVIDIA GTX 1080TI cards with Tensorflow and the Keras developed by the Google research team~[51].

We followed the method proposed in~[53] and used a transfer learning mechanism to \mbox{fine-tune} CNN models.
In~[53], Wu et al. studied the transferability of features of each layer in deep learning, and proved that transferable learning has a very good effect. 
This study provides a transfer learning idea for DCNNs composed of stacked \mbox{multi-layer} networks. 
%With the wide application of transfer learning in deep CNNs, pre-training and fine-tuning are now a very popular method for training deep learning models.

\subsection{Evaluation metrics}
\noindent
%In the object detection, the model sets a large number of prior boxes for each image in advance. 
During the training of the detection models, we updated the weight parameters according to the degree of overlap between the prediction areas and labeled areas, that is, the size of the intersection over the union (IOU). 
The IOU value can be set to different sizes; here, we set it to 0.5. 
When the IOU between the prediction areas and true areas is greater than 0.5, the prediction box is considered a true prediction and positive sample; otherwise, it is a negative sample. 
In the evaluation stage of this research, the value of the IOU was set to 0.5 to evaluate the model performance.
Precision is defined as the number ratio between positive samples and recognized samples. 
The recall is used for a certain class of objects. 
It is defined as the proportion of correctly recognized objects to the total number of such objects in the test data set. 
Precision and recall are often contradictory and not sufficient to measure the model performance alone.
 
Average precision (AP) represents the performance of the model on a certain category of objects, and the value is equal to the area enclosed by the \mbox{precision–recall} \mbox{(P-R)} curve and coordinate axis.
A \mbox{P-R} curve is simply a graph with precision values on the \mbox{y-axis} and recall values on the \mbox{x-axis}. 
In other words, the \mbox{P-R} curve contains $TP/(TP+FN)$ on the \mbox{y-axis} and $TP/(TP+FP)$ on the x-axis.
True positive (TP) means that the positive sample was detected correctly. 
False negative (FN) means that the negative sample was detected wrongly, and the false positive (FP) means that positive sample was detected wrongly. 

%AP represents a certain category of precision, which is not sufficient to measure the comprehensive performance of the model. 
The mean value of each category of AP (mAP) represents the average precision of each category of a model in the dataset, which can well reflect the comprehensive performance of the model. 
The above definition formula is as follows:
\begin{align}%
R&=\frac{TP}{TP+FN}\label{eq:recall}\\%
P&=\frac{TP}{TP+FP}\label{eq:precision}\\%
AP&=\int_{0}^{1}PRdr\label{eq:ap}\\%
mAP&=\frac{\sum_{q=1}^{Q}AveP(q)}{Q}\label{eq:mAP}%
\end{align}%
where $P$ is the abbreviation for precision, $R$ is the abbreviation for recall, and $Q$ represents the number of types of objects.

\subsection{Transfer learning based on ImageNet}
\noindent
ImageNet~[23] is widely known as a benchmark for network performance evaluation. 
It contains more than 1.2 million images and classifies large numbers of images into 1,000 categories. 
ImageNet provides \mbox{pre-trained} weights for many networks. 
Although the ImageNet dataset contains non-garbage images, the \mbox{pre-trained} weight with ImageNet can also be used in the current classification and provide the edge, corner, texture and other features of the natural image. 
These features are the basis of all the visual tasks.

Before pre-training, the final full connection layer of the convolutional classification network should be adjusted according to different tasks. 
Taking VGG16 as an example, when its \mbox{pre-training} weight is ImageNet, the output dimension of the last full connection layer of VGG16 should be correspondently modified to 30, which corresponds to the number of object classes.

\begin{figure}[tb]%
	\centering%
	\strut\hfill\strut%
	\subfigure{%
		\includegraphics[width=3.3in]{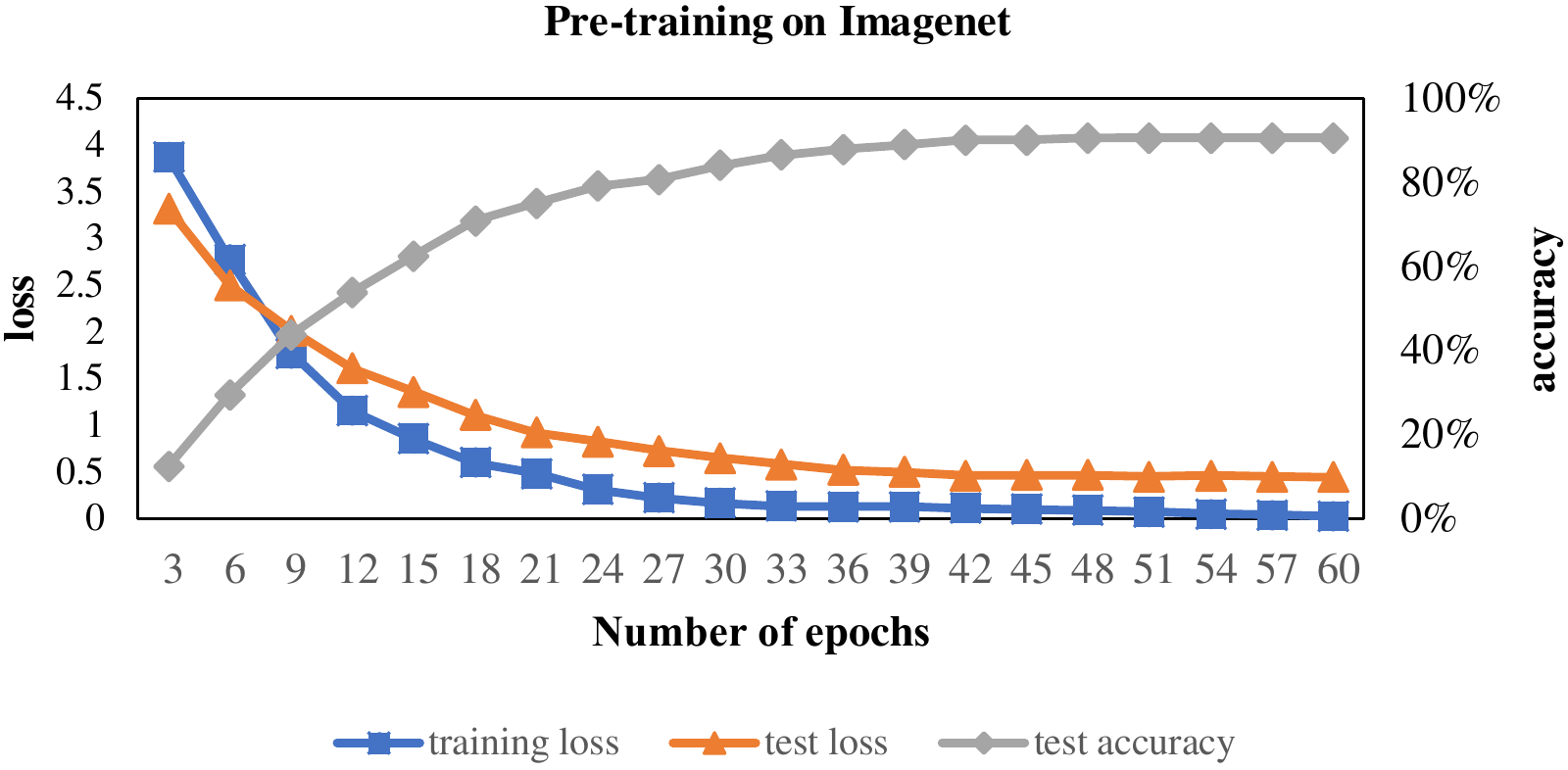}%
	}%
	%\\
	\strut\hfill\strut%
	\subfigure{%
		\includegraphics[width=3.3in]{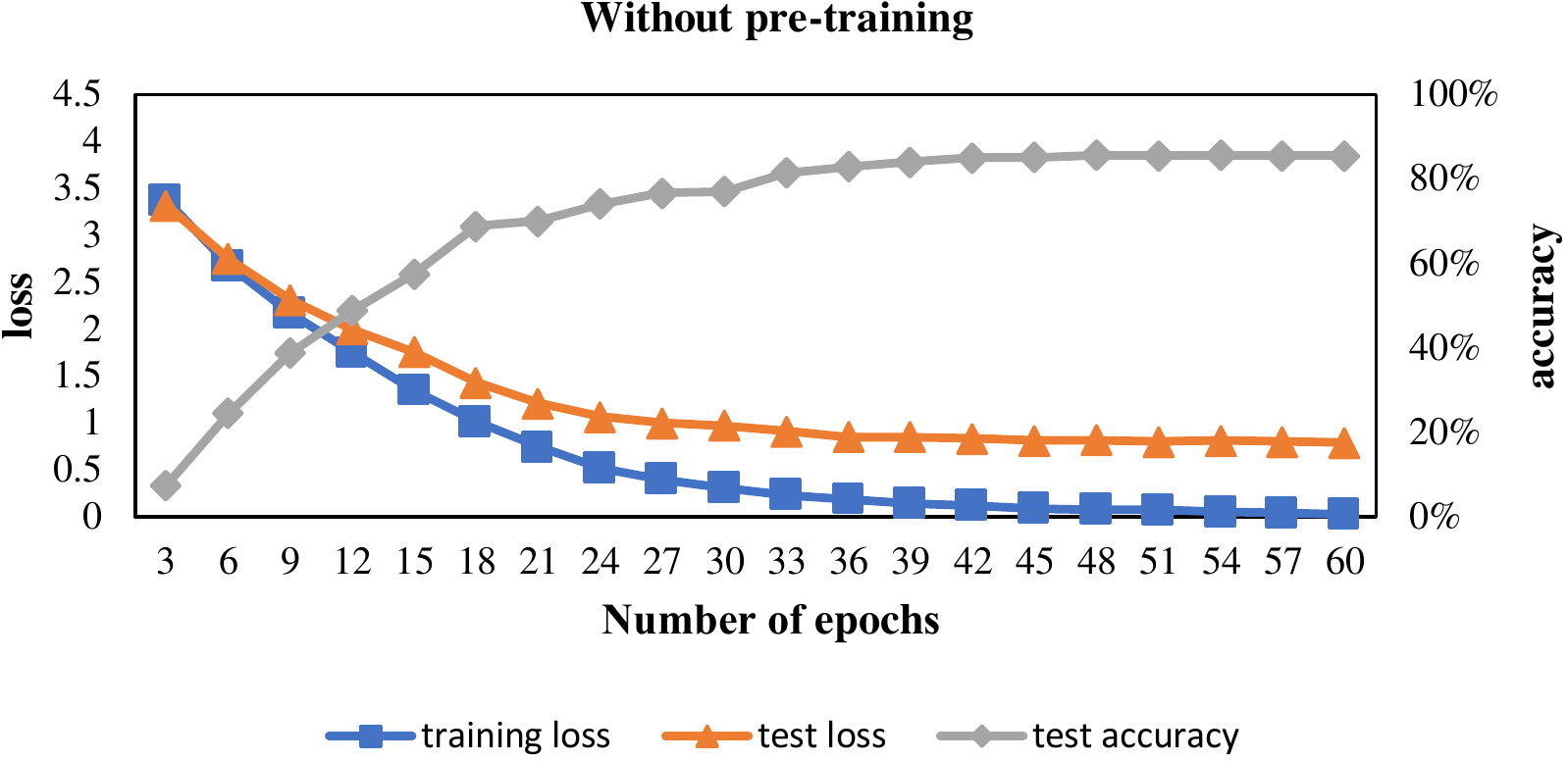}%
	}%
	%\strut\hfill\strut%
	%
	\centering
	\caption{Comparison of the loss and accuracy values with or without pre-training.}%
	\label{fig:fig-4}%
\end{figure}

In the classification, network weights based on the ImageNet dataset are taken as the initialized weights. 
Figure~\ref{fig:fig-4} shows the influence of using and not using pre-trained weights on ImageNet. 
Compared with not using a \mbox{pre-trained} network, the model accuracy and convergence speed can be improved significantly.

\subsection{Experiments of different classifications on the unlabeled HGI-30}
\noindent
We examined the performance of six \mbox{state-of-the-art} classification algorithms based on DCNNs on unlabeled \mbox{HGI-30}, namely, VGG16, MobileNet, Resnet50, DenseNet, Xception, and EfficientDet. 
For a better comparison, we also added two handcrafted \mbox{feature-based} approaches, SIFT+BOVW and HOG+SVM. 
The unlabeled \mbox{HGI-30} was divided into two parts at a ratio of $4:1$, i.e., 80\% were used for training and the rest for validation. 
The six networks used the \mbox{pre-trained} weights provided on ImageNet. 
To fairly compare the network performance, the image was uniformly adjusted to 512$\times$512 pixels before feeding it into the network. 
The initial learning rate was set to 0.00005. 
The momentum of the stochastic gradient descent was set to 0.8. 
The batch size for each iteration was set to 32. 
For SIFT+BOVW and HOG+SVM, we extracted the descriptors using a 32$\times$32 fixed-size grid with a step size of 12 pixel spacing. 
The experimental results of the above methods are shown in Table~\ref{tab:comparsion}. 
\begin{table}[tb]
	\centering
	\caption{Classification of different CNNs on unlabeled HGI-30 with or without augmentation.\\}%
	\label{tab:comparsion}
	\begin{tabular}{l|l|l}
		\hline
		\textbf{Method} & \textbf{Accuracy (\%)} & \textbf{Data Aug}\\
		\hline
		SIFT+BoVW&	64.3&	-\\
		SIFT+BoVW&	65.7&	+\\
		HOG+SVM&	67.4&	-\\
		HOG+SVM&	68.8&	+\\
		MobileNet&	85.6&	-\\
		MobileNet&	88.9&	+\\
		VGG16&	87.8&	-\\
		VGG16&	90.3&	+\\
		ReseNet50&	89.2&	-\\
		ReseNet50&	91.7&	+\\
		DenseNet&	92.1&	+\\
		XCeption&	92.6&	+\\
		EfficientDet&	93.2&	+	\\	
		\hline
	\end{tabular}
\end{table}

EfficientDet achieved the best effect with an accuracy of 93.2\%. 
Benefiting from the recombination coefficient to balance the width, depth, and resolution of the network, EfficientDet simultaneously extended them and performed excellently. 
The performance gap among ResNet, DenseNet, and Xception is very small compared to that of EfficientDet. 
Although MobileNet's accuracy is not the best, the number of parameters is minimal. 
Particularly, DCNN algorithms have excellent performance, far better than those of traditional algorithms, where transfer learning plays an important role. 
We also compared the sensitivity of traditional and DCNN algorithms to data augmentation. 
The results show that DCNN algorithms are highly sensitive to data augmentation, whereas traditional algorithms are not.

\subsection{Experiments of different detections on the labeled HGI-30}
\noindent
To evaluate the detection performance of DCNNs, six \mbox{state-of-the-art} detection methods were investigated on \mbox{HGI-30}: SSD, YOLOv3, YOLOv4, Faster RCNN, M2Det, and EfficientDet. 
In this experiment, the \mbox{pre-training} of the five detection models was provided by the PASCAL VOC dataset~[52]. 
The \mbox{HGI-30} dataset was divided into two parts in a ratio of $4:1$, with 80\% used for training and the rest for validation.

Table~\ref{tab:comparision-detection} provides the detection results on the \mbox{HGI-30} dataset in comparison with six \mbox{state-of-the-art} methods.
Overall, YOLOv4 achieved the best effect. 
It has introduced numerous techniques and tools, including CSPDarknet53 as a backbone, Mosaic data enhancement, and improved SAM. 
SSD has the worst comprehensive performance. 
EfficientDet has achieved a very effective performance by balancing the width, depth and resolution and improving the \mbox{bi-directional} feature pyramid network. 
Particularly, EfficientDet has a fairly high training hardware requirement.
\begin{table}[tb]%
	\centering%
	\setlength{\belowcaptionskip}{0.2cm}
	\caption{Detection average precision (\%) of alternative methods on \mbox{HGI-30}}%
	\label{tab:comparision-detection}%
	\resizebox{0.99\linewidth}{!}{%
		\begin{tabular}{l|rrrrrr}
			\hline
			\diagbox{Class}{Method}&EfficientDet& Faster RCNN&M2Det &SSD&  YOLOv3&YOLOv4\\
			\hline
			applecore&	0.9512&	0.9254&	0.9413&	0.9017&	0.9163&	\textbf{0.9692}\\
			banana peel&	\textbf{0.7438}&	0.6743&	0.7093&	0.6535&	0.6675&	\textbf{0.7438}\\
			battery	&\textbf{0.7631}&	0.7057	&0.7405&	0.6763&	0.7198&	0.7592\\
			book	&0.7894&	0.7513&	0.7581&	0.7257&	0.7616&	\textbf{0.8597}\\
			buttonbattery	&0.5319	&0.4917	&0.4692	&0.5182	&0.4673&	\textbf{0.5425}\\
			can	&0.7948	&0.7807&	0.8432&	0.7738&	0.7914&	\textbf{0.8697}\\
			capsule	&0.7650	&0.7121	&0.7359	&0.7246	&0.6873&	\textbf{0.8168}\\
			carton	&0.8596	&\textbf{0.8731}&	0.8404&	0.8492&	0.8496&	0.8622\\
			cigarette butt	&\textbf{0.8635}	&0.8236	&0.8359	&0.8255	&0.7929&	0.8145\\
			cigarette case	&0.6723&	0.6219&	0.6077&	0.5868&	0.6453&	\textbf{0.6926}\\
			glass bottle&	0.8049	&0.7862	&0.8303	&0.8101	&0.8062	&\textbf{0.8436}\\
			lunch box	&\textbf{0.5989}&	0.5786&	0.5483&	0.5394&	0.5608&	0.5932\\
			mask	&0.7761	&0.7634	&\textbf{0.7858}	&0.7559	&0.7647	&0.7735\\
			mobilephone	&0.8535&	0.8664&	0.8301&	0.8674&	0.8382&	\textbf{0.8835}\\
			modulator tube	&0.5622	&0.5416	&0.5517	&0.5138	&0.5409&	\textbf{0.6085}\\
			paper cup	&0.8988&	\textbf{0.9020}&	0.8711&	0.8866&	0.8868	&0.8944\\
			pencil	&0.8393	&0.7971	&0.8525	&0.8343	&0.8215	&\textbf{0.8783}\\
			plastic bottle&	0.7752&	0.7499&	0.7614&	0.7287&	0.7507&	\textbf{0.8186}\\
			remotecontrol	&0.8641	&0.8377	&0.8458	&0.8401	&0.8523&\textbf{0.8837}\\
			rice	&0.9677&	0.9439&	0.9446&	0.9213	&0.9262&	\textbf{0.9715}\\
			shoe	&0.8248	&0.7865	&0.8496	&0.7926&	0.8090	&\textbf{0.8697}\\
			t-shirt&	0.9141&	0.8838	&0.8979&	0.8541&	0.8695	&\textbf{0.9174}\\
			tea leaf	&\textbf{0.8597}	&0.8152	&0.8262	&0.7961	&0.8091	&0.8272\\
			thermometer	&0.5301&	0.5013&	0.5293&	0.5196&	0.4904&	\textbf{0.5681}\\
			tin can	& 0.8448& 0.8263&	0.8456&	0.8219	&0.7901&	\textbf{0.8458}\\
			toothbrush	&\textbf{0.8148}	&0.7616&	0.7912	&0.7512	&0.7643&	0.7785\\
			trousers	&0.8340	&0.8415&	0.8507	&0.8172&	0.8307&	\textbf{0.8617}\\
			vegetable leaf	&0.6207	&0.6321	&0.6226	&0.5886	&0.6174	&\textbf{0.6543}\\
			waste paper	&\textbf{0.7249}&	0.6528&	0.7149&	0.6157&	0.6541&	0.6632\\
			watermelon peel	&0.6435	&0.6147	&0.6324	&0.5763&	0.5938&\textbf{0.6714}\\
			\hline
			\emph{mAP} & 0.7762  &0.7480  &0.7621 &0.7355  &0.7425	&\textbf{0.7912}\\
			\hline
	\end{tabular}}%
\end{table}
%
%We compare the performance of the six object detection models with and without data augmentation, and carry out quantitative analysis on the performance. 
%The mAP of the six state-of-the-art methods is illustrated in Figure~\ref{fig:fig-5}.
%
%\begin{figure}[tb]%
%	\centering%
%	\includegraphics[width=0.477\textwidth]{fig-map.pdf}%
%	\caption{mAP of the six state-of-the-art methods}%
%	\label{fig:fig-5}%
%\end{figure}%
%

\begin{figure}[tb]%
	\centering%
	\includegraphics[width=0.477\textwidth]{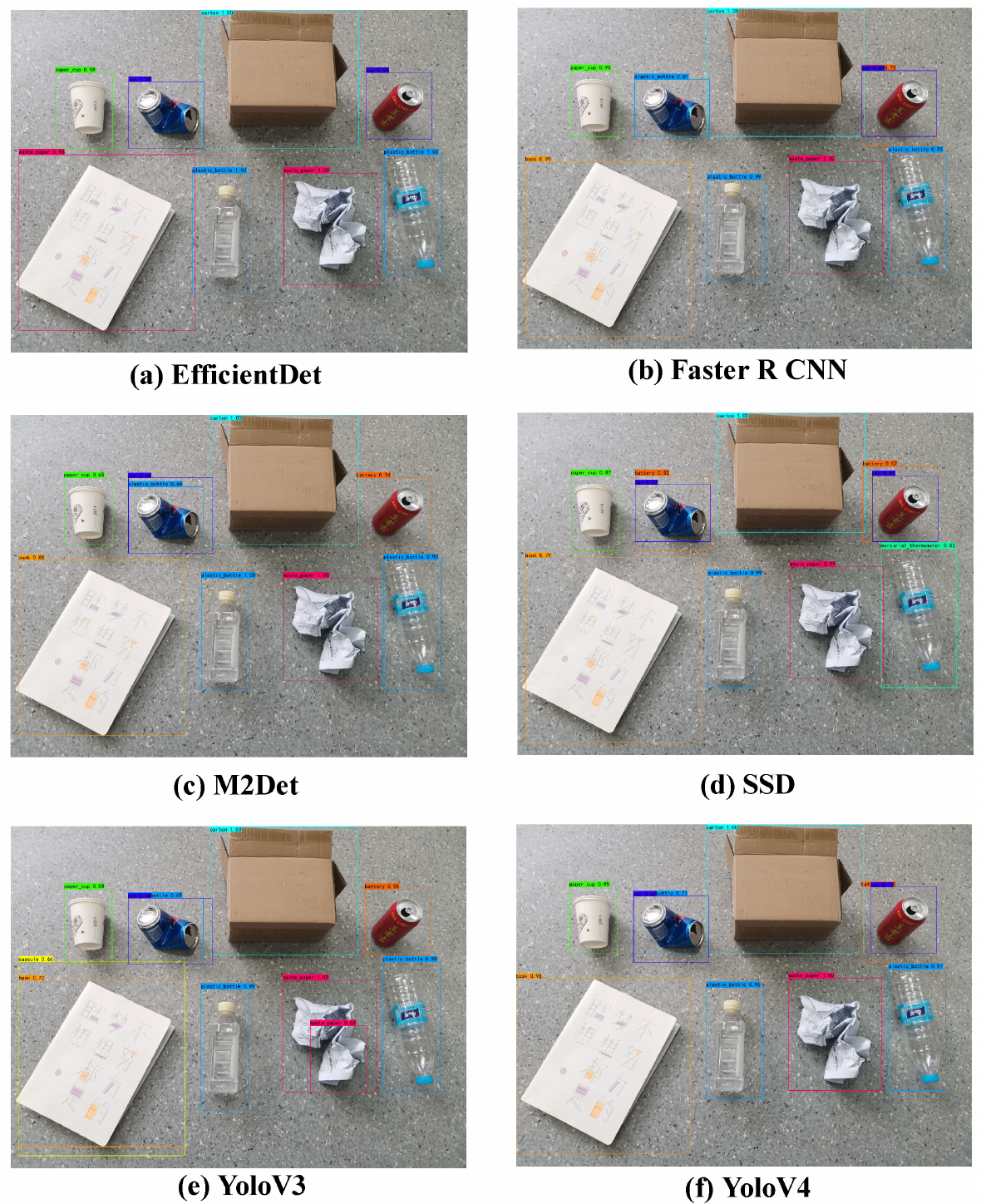}%
	\caption{Examples of the detection results by the \mbox{state-of-the-art} methods}%
	\label{fig:fig-6}%
\end{figure}

As an outstanding representative of the two stage algorithms, YOLOv3 also achieves a good result; but compared with M2Det, there are still some gaps. 
As shown in Table~\ref{tab:comparision-detection}, the average mAP of the six target detection algorithms in this dataset is 76\%.
Hence, we can conclude that HGI-30 is a challenge for object detection.  
In Figure~\ref{fig:fig-6}, we provide examples of detection results using the six algorithms.

\begin{figure*}[tb]%
	\centering%
	\includegraphics[width=0.90\textwidth]{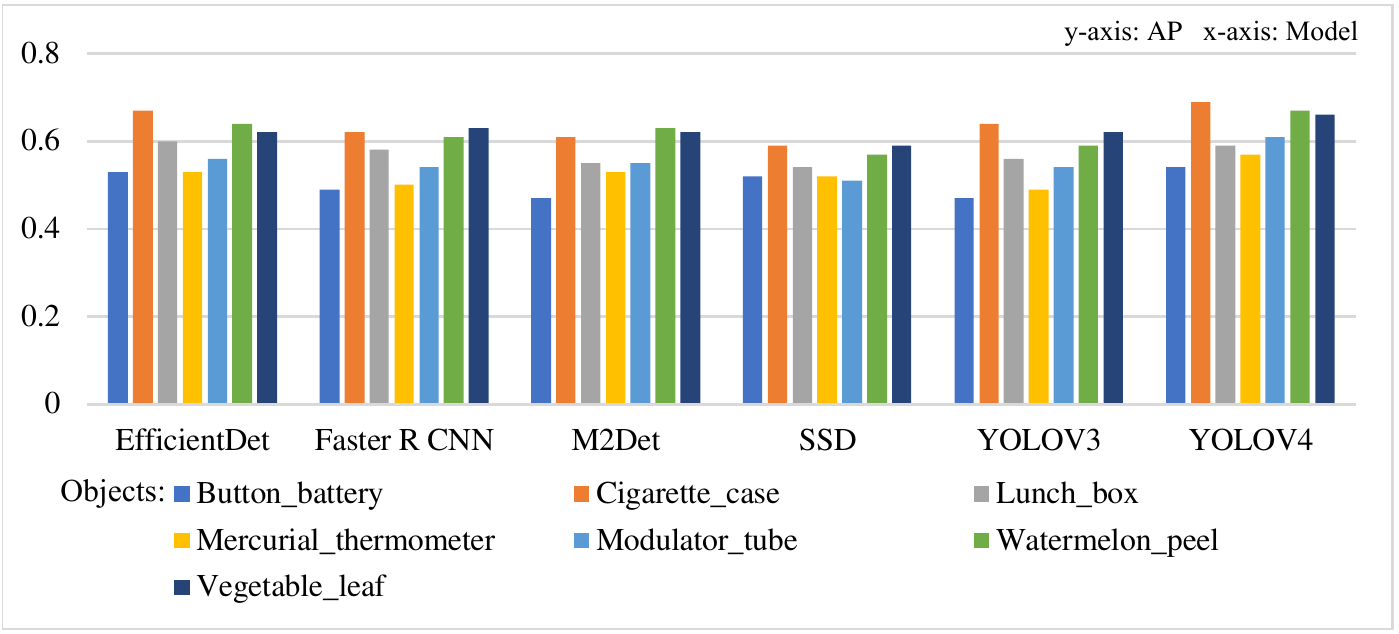}%
	\caption{Comparison of the detection results on small objects or irregularly shaped objects}%
	\label{fig:fig-7}%
\end{figure*}

We also validated several types of typical garbage on six models and found that none of them are satisfactory for recognizing small objects or irregularly shaped objects. 
Although the general performance of SSD is the worst, it has a good effect on the detection of small targets, because the expansion convolution is added. 
Here, YOLO still has the best performance.
In addition, the DCNNs are more sensitive to texture, compared to the irregularly shaped or small garbage. 
Objects with relatively regular textures and fixed geometry also tend to have good detection accuracy. 
The results are shown in Figure~\ref{fig:fig-7}.

\section{Conclusion and Future Work}
\noindent
In this study, we first analyzed the characteristics of garbage and the current status of garbage classification. 
We then reviewed the state of the art in this domain and found that, as a consequence of the lack of suitable benchmarks, most approaches were evaluated on different datasets under different experimental settings. 
Thus, it is hard to fairly compare the results published in the literature. 
To solve this problem, we make the \mbox{HGI-30} dataset available in the online repository~[15].

Next, we performed experiments on classification and detection on this dataset and analyzed the characteristics of current mainstream networks. 
Experiments showed that \mbox{HGI-30} can reflect DCNNs' characteristics, and garbage image recognition is a great challenge for existing object recognition algorithms based on DCNNs. 
Finally, we present the acquisition path of \mbox{HGI-30}.

In future work, we will expand the categories and quantities of garbage in the dataset, and run more algorithms. 
In addition, we will develop better models on this dataset to facilitate the development of target recognition algorithms and environmental protection.
\vskip 2mm
\large
\noindent
\textbf{Acknowledgment}
\vskip 2mm

\Acknow
\noindent
%We acknowledge support from the Youth Project of the Provincial Natural Science Foundation of Anhui 1908085QF285, 1908085QF262, 1908085MF184 and 1908085MF185, the Key Research Plan of Anhui 202104d07020006, the University Natural Sciences Research Project of Anhui Province KJ2020A0661 and KJ2019ZD61, China's Post-doctoral Science Fund~2020M681989, the National Natural Science Foundation of China under grant 62176085.
We acknowledge support from the Youth Project of the Provincial Natural Science Foundation of Anhui 1908085QF285, 1908085QF262, and 1908085MF184, the Key Research Plan of Anhui 202104d07020006, the University Natural Sciences Research Project of Anhui Province KJ2020A0661 and KJ2019A0834, the National Natural Science Foundation of China under grant 61906062 and 62176085.

%\vskip 2mm
%\zihao{5}
%\noindent
%\textbf{\zihao{5}References}
%\vskip 2mm

\renewcommand\refname{\large
\mbox{}
\clearpage
\clearpage
\large
\noindent

\textbf{References}}
\begin{thebibliography}{99}
	\normalsize \addtolength{\itemsep}{-1em}
	\vspace {1.5mm}

\bibitem[1]{1}
Zhaojie. D, Chenjie. Z, Jiajie. W, Yifan. Q, and Gang. C, Garbage Classification System Based on AI and IoT presented at the 15th IEEE International Conference on Computer Science \& Education, Delft, Netherlands, 2020. %\textbf{(Paper presented at conferences style)}

\bibitem[2]{2}
Zhang Q X, Lin G H, Zhang Y M, Xu G, and Wang J J, Wildland forest fire smoke detection based on faster R-CNN using synthetic smoke images, \emph{Procedia engineering,} vol. 221, no. 3, pp.441-446, Mar. 2018.

\bibitem[3]{3}
Chen. H, Chen. A, Xu. L, Xie. H, Qiao. H, Lin. Q, and Cai. K, A deep learning CNN architecture applied in smart near-infrared analysis of water pollution for agricultural irrigation resources, \emph{Agricultural Water Management}, doi:10.1016/j.agwat.2020.106303. %\textbf{(Article by DOI style)}

\bibitem[4]{4}
Han. C, Li. G, Ding. Y, Yan. F, and Bai. L, Chimney Detection Based on Faster R-CNN and Spatial Analysis Methods in High Resolution Remote Sensing Images, \emph{Sensors}, doi:10.3390/s20164353. %\textbf{(Article by DOI style)}

\bibitem[5]{5}
Hu. G, Wang. H, Zhang. Y, and Wan. M, Detection and severity analysis of tea leaf blight based on deep learning, \emph{Computers \& Electrical Engineering}, doi: 10.1016/j.compelceng.2021.107023. %\textbf{(Article by DOI style)}

\bibitem[6]{6}
Datta. D, and Jamalmohammed. S. B, Detection and severity analysis of tea leaf blight based on deep learning, \emph{Applications of Artificial Intelligence for Smart Technology}, doi: 10.4018/978-1-7998-3335-2.ch016. %\textbf{(Article by DOI style)}

\bibitem[7]{7}
Zeng. D, Zhang. S, Chen. F, and Wang. Y, Detection and severity analysis of tea leaf blight based on deep learning, \emph{IEEE Access}, doi: 10.1109/ACCESS.2019.2932117. %\textbf{(Article by DOI style)}

\bibitem[8]{8}
Ye. A, Pang. B, Jin. Y, and Cui. J, A YOLO-based Neural Network with VAE for Intelligent Garbage Detection and Classification presented at the 3rd International Conference on Algorithms, Computing and Artificial Intelligence, 2020. %\textbf{(Paper presented at conferences style)}

\bibitem[9]{9}
Nie. Z, Duan. W, and Li. X, Domestic garbage recognition and detection based on Faster R-CNN, \emph{Journal of Physics: Conference Series}, doi:10.1088/1742-6596/1738/1/012089. %\textbf{(Article by DOI style)}

\bibitem[10]{10}
Bai. J, Lian. S, Liu. Z, Deep learning based robot for automatically picking up garbage on the grass, \emph{IEEE Transactions on Consumer Electronics}, doi:10.1109/TCE.2018.2859629. %\textbf{(Article by DOI style)}

\bibitem[11]{11}
Huiyu. L, O. O. G, and Kim. S. H, Automatic Classifications and Recognition for Recycled Garbage by Utilizing Deep Learning Technology presented at the 7th International Conference on Information Technology: IoT and Smart City, 2019. %\textbf{(Paper presented at conferences style)}

\bibitem[12]{12}
Mittal. G, Yagnik. K. B, Garg. M, and Krishnan. N. C, Spotgarbage: smartphone app to detect garbage using deep learning presented at the ACM International Joint Conference on Pervasive and Ubiquitous Computing, Heidelberg, Germany, 2016. %\textbf{(Paper presented at conferences style)}

\bibitem[13]{13}
Jia. G, Zhu. Y, Han. G, Chan. S, and Shu. L, STC: an intelligent trash can system based on both NB-IoT and edge computing for smart cities, \emph{Enterprise Information Systems} vol. 14, no. 9-10, pp.1422-1438, Sep. 2020. %\textbf{(Journal style)}

\bibitem[14]{14}
Rabano. S. L, Cabatuan. M. K, Sybingco. E, Dadios. E. P, and Calilung. E. J, Common garbage classification using mobilenet presented at the 10th IEEE International Conference on Humanoid, Nanotechnology, Information Technology, Baguio City, Philippines, 2018. %\textbf{(Paper presented at conferences style)}

\bibitem[15]{15}
Li. H. Y, HGI-30 DATA Set [Data set], http://doi.org/10.5281/zenodo.4646699, 2021, Mar. 31. %\textbf{(Online document style)}
%\end{thebibliography}

\bibitem[16]{16}
Ojala. T, Pietikainen. M, and Maenpaa. T, Multiresolution gray-scale and rotation invariant texture classification with local binary patterns, \emph{IEEE Transactions on pattern analysis and machine intelligence} vol. 24, no. 7, pp.971-987, Jul. 2002. %\textbf{(Journal style)}

\bibitem[17]{17}
Zhang. L, Chu. R, Xiang. S, Liao. S, and Li. S. Z, Face detection based on multi-block lbp representation presented at the International conference on biometrics, 2007. %\textbf{(Paper presented at conferences style)}
\bibitem[18]{18}
Lowe. D. G, Distinctive image features from scale-invariant keypoints, \emph{International journal of computer vision} vol. 60, no. 2, pp.91-110, Nov. 2004. %\textbf{(Journal style)}

\bibitem[19]{19}
Tao. C, Tan. Y, Cai. H, and Tian. J, Airport detection from large IKONOS images using clustered SIFT keypoints and region information, \emph{IEEE Geoscience and Remote Sensing Letters} vol. 8, no. 1, pp.128-132, Jul. 2010. %\textbf{(Journal style)}

\bibitem[20]{20}
Dalal. N, and Triggs. B, Histograms of oriented gradients for human detection presented at the 22th IEEE Conference on Computer Vision and Pattern Recognition, San Diego, USA 2005. %\textbf{(Paper presented at conferences style)}

\bibitem[21]{21}
Pang. Y, Yuan. Y, Li. X, and Pan. J, Efficient HOG human detection. Signal Processing, \emph{Signal Processing}, doi:10.1016/j.sigpro.2010.08.010. %\textbf{(Article by DOI style)}

\bibitem[22]{22}
Krizhevsky. A, Sutskever. I, Hinton. G. E, Imagenet classification with deep convolutional neural networks, \emph{Advances in Neural Information Processing Systems}, doi:10.1145/3065386. %\textbf{(Article by DOI style)}

\bibitem[23]{23}
Deng. J, Dong. W, Socher. R. Li, L. J, Li. K, and Fei-Fei. L, Imagenet: A large-scale hierarchical image database presented at the 26th IEEE Conference on Computer Vision and Pattern Recognition, Miami, USA, 2009. %\textbf{(Paper presented at conferences style)}

\bibitem[24]{24}
Simonyan. K, Zisserman. A, Very deep convolutional networks for large-scale image recognition, https://arxiv.org/abs/1409.1556, 2015, Apr. 10. %\textbf{(Online document style)}
%\end{thebibliography}

\bibitem[25]{25}
Szegedy. C, Liu. W, Jia. Y, Sermanet. P, Reed. S, Anguelov. D, and Rabinovich, Going deeper with convolutions presented at the 32th IEEE Conference on Computer Vision and Pattern Recognition, Boston, USA, 2015. %\textbf{(Paper presented at conferences style)}

\bibitem[26]{26}
He. K, Zhang. X, Ren. S, and Sun. J, Deep residual learning for image recognition presented at the IEEE Conference on Computer Vision and Pattern Recognition, Las Vegas, USA, 2016. %\textbf{(Paper presented at conferences style)}

\bibitem[27]{27}
Hu. J, Shen. L, and Sun. G, Squeeze-and-excitation networks presented at the 35th IEEE Conference on Computer Vision and Pattern Recognition, Salt Lake City, USA, 2018. %\textbf{(Paper presented at conferences style)}

\bibitem[28]{28}
Liu. Y. H, Feature extraction and image recognition with convolutional neural networks, \emph{Journal of Physics: Conference Series}, doi:10.1088/1742-6596/1087/6/062032. %\textbf{(Article by DOI style)}

\bibitem[29]{29}
Jogin. M, Madhulika, M. S, Divya, G. D, Meghana, R. K, and Apoorva. S, Feature extraction using convolution neural networks (CNN) and deep learning presented at the 3rd IEEE International Conference on Recent Trends in Electronics, Information \& Communication Technology, Bangalore, India, 2018. %\textbf{(Paper presented at conferences style)}

\bibitem[30]{30}
Shorten. C, Khoshgoftaar. T. M, A survey on image data augmentation for deep learning, \emph{Journal of Big Data}, doi:10.1186/s40537-019-0197-0. %\textbf{(Article by DOI style)}   

\bibitem[31]{31}
Zhang. R. S, Quan. W. Z, Fan. L. B, Hu. L. M, and Yan. D. M, Distinguishing computer-generated images from natural images using channel and pixel correlation, \emph{Journal of Computer Science and Technology}, doi:10.1007/s11390-020-0216-9. %\textbf{(Article by DOI style)}

\bibitem[32]{32}
Zhang. X. J, Lu. Y. F, and Zhang. S. H, Multi-task learning for food identification and analysis with deep convolutional neural networks, \emph{Journal of Computer Science and Technology}, doi:10.1007/s11390-016-1642-6. %\textbf{(Article by DOI style)}

\bibitem[33]{33}
Jia. J. G, Zhou. Y. F, Hao. X. W, Li. F, Desrosiers. C, and Zhang. C. M, Two-Stream Temporal Convolutional Networks for Skeleton-Based Human Action Recognition, \emph{Journal of Computer Science and Technology}, doi:10.1007/s11390-020-0405-6. %\textbf{(Article by DOI style)}

\bibitem[34]{34}
Minaee. S, Boykov. Y. Y, Porikli. F, Plaza. A. J, Kehtarnavaz. N, and Terzopoulos. D, Image segmentation using deep learning: A survey, \emph{IEEE Transactions on Pattern Analysis and Machine Intelligence}, doi:10.1109/TPAMI.2021.3059968. %\textbf{(Article by DOI style)}

\bibitem[35]{35}
Wu. Y, Lim. J, Yang. M. H, Online object tracking: A benchmark presented at the 30th IEEE Conference on Computer Vision and Pattern Recognition, Portland, America, 2013. %\textbf{(Paper presented at conferences style)}

\bibitem[36]{36}
Xie. Z. F, Guo. Y. C, Zhang. S. H, Zhang. W. J, and Ma. L. Z, Multi-exposure motion estimation based on deep convolutional networks, \emph{Journal of Computer Science and Technology}, doi:10.1007/s11390-018-1833-4. %\textbf{(Article by DOI style)}

\bibitem[37]{37}
Caroppo. A, Leone. A, and Siciliano. P, Comparison Between Deep Learning Models and Traditional Machine Learning Approaches for Facial Expression Recognition in Ageing Adults, \emph{Journal of Computer Science and Technology}, doi:10.1007/s11390-020-9665-4. %\textbf{(Article by DOI style)}

\bibitem[38]{38}
Wang. P, Fan. E, and Wang. P, Comparative analysis of image classification algorithms based on traditional machine learning and deep learning, \emph{Pattern Recognition Letters}, doi:10.1016/j.patrec.2020.07.042. %\textbf{(Article by DOI style)}

\bibitem[39]{39}
Ren. S, He. K, Girshick. R, and Sun. J, Faster r-cnn: Towards real-time object detection with region proposal networks, http://arxiv.org/abs/1506.01497, 2016, Jan. 6. %\textbf{(Online document style)}
%\end{thebibliography}

\bibitem[40]{40}
He. K, Gkioxari. G, Dollár. P, and Girshick. R, Mask r-cnn presented at the 34th IEEE Conference on Computer Vision and Pattern Recognition, Venice, Italy, 2017. %\textbf{(Paper presented at conferences style)}

\bibitem[41]{41}
Dai. J, Qi. H, Xiong. Y, Li. Y, Zhang. G, Hu. H, and Wei. Y, Deformable convolutional networks presented at the 34th IEEE Conference on Computer Vision and Pattern Recognition, Venice, Italy, 2017. %\textbf{(Paper presented at conferences style)}

\bibitem[42]{42}
Nie. Z, Duan. W, and Li X, Domestic garbage recognition and detection based on Faster R-CNN, \emph{Journal of Physics: Conference Series}, doi:10.1088/1742-6596/1738/1/012089. %\textbf{(Article by DOI style)}

\bibitem[43]{43}
Liu. W, Anguelov. D, Erhan. D, Szegedy. C, Reed. S, Fu. C. Y, and Berg. A. C, Ssd: Single shot multibox detector presented at the 14th European Conference on Computer Vision, Amsterdam, The Netherlands, 2016. %\textbf{(Paper presented at conferences style)}
\bibitem[44]{44}
Redmon. J, Divvala. S, Girshick. R, and Farhadi. A, You only look once: Unified, real-time object detection presented at the 33th IEEE Conference on Computer Vision and Pattern Recognition, Las Vegas, USA, 2016. %\textbf{(Paper presented at conferences style)}

\bibitem[45]{45}
Redmon. J, and Farhadi. A, YOLO9000: better, faster, stronger presented at the 34th IEEE Conference on Computer Vision and Pattern Recognition, Hawaii, USA, 2017. %\textbf{(Paper presented at conferences style)}

\bibitem[46]{46}
Redmon. J, Farhadi. A, Yolov3: An incremental improvement, https://arxiv.org/abs/1804.02767, 2018, Apr. 8. %\textbf{(Online document style)}
%\end{thebibliography}

\bibitem[47]{47}
Zhao. Q, Sheng. T, Wang. Y, Tang. Z, Chen. Y, Cai. L, and Ling. H, M2det: A single-shot object detector based on multi-level feature pyramid network presented at the 31th AAAI Conference on Artificial Intelligence, Hawaii, USA, 2017. %\textbf{(Paper presented at conferences style)}

\bibitem[48]{48}
Tan. M, Pang. R, and Le. Q. V, Efficientdet: Scalable and efficient object detection presented at the 37th IEEE Conference on Computer Vision and Pattern Recognition, Seattle, USA, 2020. 
%\textbf{(Paper presented at conferences style)}

\bibitem[49]{49}
Chen. Q, Garbage Classification Detection Based on Improved YOLOV4, \emph{Journal of Computer and Communications}, doi:10.4236/jcc.2020.812023. 
%\textbf{(Article by DOI style)}

\bibitem[50]{50}
Islam. M. A, Jia. S, and Bruce. N. D, How much position information do convolutional neural networks encode?, https://arxiv.org/abs/2001.08248, 2020, Jan. 22. %\textbf{(Online document style)}
%\end{thebibliography}

\bibitem[51]{51}
Abadi, Martín, et al., Tensorflow: A system for large-scale machine learning presented at the 12th {USENIX} Symposium on Operating Systems Design and Implementation, Savannah, USA, 2016. 

%\bibitem[52]{52}
%Yosinski. J, Clune. J, Bengio. Y, and Lipson. H, How transferable are features in deep neural networks?, https://arxiv.org/abs/1411.1792, 2014, Nov. 6. %\textbf{(Online document style)}
%\end{thebibliography}

\bibitem[52]{52}
Everingham. M, Van. Gool. L, Williams. C. K, Winn. J, and Zisserman. A, The pascal visual object classes (voc) challenge, \emph{International Journal of Computer Vision}, doi:10.1007/s11263-009-0275-4. %\textbf{(Article by DOI style)}

\bibitem[53]{53}
Wu, Z.Z., Wan, S.H., Wang, X.F., Tan, M., Zou, L., Li, X.L. and Chen, Y.,  A benchmark data set for aircraft type recognition from remote sensing images. \emph{Applied Soft Computing}, doi:10.1016/j.asoc.2020.106132.

\end{thebibliography}
\end{document}